\documentclass[runningheads]{llncs}
\usepackage[T1]{fontenc}
\usepackage{xcolor}
\usepackage{graphicx}
\usepackage{appendix}
\begin{document}
\title{Valuing Players Over Time}
%
%
\author{Tiago Mendes-Neves\inst{1,2}\orcidID{0000-0002-4802-7558} \and\\
Luís Meireles\inst{1,3}\orcidID{0000-0003-3194-8139} \and\\
João Mendes-Moreira\inst{1,2}\orcidID{0000-0002-2471-2833}}
\authorrunning{Tiago Mendes-Neves et al.}
%
\institute{Faculdade de Engenharia, Universidade do Porto, Porto, Portugal \and
LIAAD - INESC TEC, Porto, Portugal
\and
FC Porto – Futebol SAD, Porto, Portugal\\
\email{tmendesnvs@gmail.com}}
\maketitle              
\begin{abstract}
In soccer (or association football), players quickly go from heroes to zeroes, or vice-versa. Performance is not a static measure but a somewhat volatile one. Analyzing performance as a time series rather than a stationary point in time is crucial to making better decisions. This paper introduces and explores I-VAEP and O-VAEP models to evaluate actions and rate players' intention and execution. Then, we analyze these ratings over time and propose use cases to fundament our option of treating player ratings as a continuous problem. As a result, we present who were the best players and how their performance evolved, define volatility metrics to measure a player's consistency, and build a player development curve to assist decision-making.

\end{abstract}
\section{Introduction} \label{sec:Introduction}
Measuring the quality of soccer players is the main task of many staff members at soccer clubs. Coaches, scouts and sporting directors evaluate players daily, with the goal of helping tactical or recruitment decisions. Judgments about player quality significantly impact a team's chances of success (or failure).

There are many approaches to rate players’ quality. Traditionally, soccer clubs work with qualitative analysis from scouts, generating reports from live and on-demand video analysis. Recently, the trend started to change towards data-driven quantitative analysis of players. Here, data analysts use tools built with event or tracking data, which summarize data to key features that the analyst uses to filter players that are adequate for their team. Teams and service providers are also investing in machine learning models, for example, to predict how much value players add to their team or to predict a player’s market value. These models also have value from the regulators' perspective, such as policing accounting fraud in soccer transfers.

Due to the value of these player rating models, even from the outsider's perspective (fan engagement, fantasy, betting), there are many approaches in the academic literature and private businesses, namely, live score apps and data services providers. Our work focuses on VAEP, a framework that evaluates actions in a game by estimating how a action changes the probability of scoring.

Instantaneous approaches to rate player quality are helpful. They summarize the player to a value (or a set of values), making it much easier to guide decision-making. Nevertheless, they have one flaw: they do not consider how the player developed over time. By focusing on a single value, decisions can be misguided. What if the recent player performance was an outlier? What if luck played an important role in recent performances? Is the player improving over time? How likely is it that the player hit a skill ceiling? From the perspective of recruitment and player development, it is important to track a player’s performance over time to make a more informed (and better) decision.




In this paper, we make the following contributions:

\begin{itemize}
    \item \textbf{Section \ref{sec:Literature Review}} reviews the literature on machine learning-based player evaluation systems and fundaments our choice regarding VAEP.
    \item \textbf{Section \ref{sec:Defining the VAEP model}} describes our VAEP models with three significant changes versus the original proposal \cite{decroos_actions_2019}: (1) we modify the label to reflect how much time before a goal the action occurs, (2) address the problem using regression, and (3) use the Random Forest algorithm with a smaller set of features which creates a simpler model. Furthermore, we distinguish between the Intent VAEP (I-VAEP) model and the Outcome-aware VAEP (O-VAEP) model.
    \item \textbf{Section \ref{sec:Creating tracking metrics}} details the process of extracting series and preprocessing them to obtain robust tracking metrics for players’ skills. Specifically, we use data from Spanish La Liga between 2009 and 2019 to generate the time series. We debate the differences between using the game date, or cumulative games played as the series index, and create short-term and long-term metrics.
    \item \textbf{Section \ref{sec:Experimental Use Cases}} presents the following experiments:
    \begin{itemize}
        \item We show visualizations of the data to find the best players and their best way of creating value for the team. We exemplify how the generated data helps us visualize playstyle changes, with an example of the evolution of Lionel Messi's playstyle over time.
        \item Continuing to explore our dataset, we evaluate whether player performance volatility should play a more critical role in decision-making.
        \item We build an expected player development curve with our time-series data, handling dataset biases such as younger players in our dataset being much better than the average young player.
    \end{itemize}
    \item \textbf{Section \ref{sec:Conclusion}} discusses the advantages and disadvantages of treating the player evaluation metrics as time series while also discussing some of the limitations of our overall approach.
\end{itemize}

\section{Literature Review} \label{sec:Literature Review}
There are many options for building player evaluation frameworks. The simplest is perhaps the ELO rating system \cite{elo_rating_1978} . Originally developed to measure the strength of chess players, ELO systems can also evaluate relative team strength. The original ELO system does not rate individual players in team sports. However, we can adapt the model to measure a player’s contribution by calculating the differences in the ELO rating when the player is on and off the pitch. This system works quite well in high-rotation sports, like basketball, where players cycle through bench time and play time several times during a game.

In the past decade, the concept of expected goals (xG) \cite{pollard_estimating_2004} gained popularity in the field. xG measure the probability of a shot ending in a goal. For this, it uses game state information, like the ball's position, which part of the body the player is using, the position of the goal keeper, and much more. The major drawback of xG is that it only evaluates one action (shots). Although useful to measure the performance of strikers, it does not scale to other positions. Expected assists, which measures the likelihood of a pass leading to an assist, has similar drawdowns.

Some frameworks enable extending player evaluation across multiple action types. Expected Threat (xT) \cite{roy_valuing_2020}, Expected Possession Value (EPV) \cite{fernandez_framework_2021}, and Valuing Actions by Estimating Probabilities (VAEP) \cite{decroos_actions_2019} are frameworks that calculate the likelihood of a position or an action leading to a goal in the near future. In a way, these methods extend similar concepts to xG to a broader set of data.

In this paper, we use a custom VAEP model to estimate the value of actions in our dataset. Different methods can be used to generate results equivalent to the ones we obtained. However, we opted for VAEP because (1) it is easier to apply to the datasets available, and (2) there is a higher familiarity of the authors with this framework.

\section{Defining the VAEP model} \label{sec:Defining the VAEP model}
To build our VAEP models, we used the following event data: the training set is from the German Bundesliga, English Premier League, French Ligue 1, and Italian Serie A from 2018-2019. The test set is from the Spanish La Liga between 2009-2010 and 2018-2019. The final feature set is available in the Appendix \ref{ap:features}.



From the feature list, there is a particular set of features of interest: \textit{endAngleToGoal}, \textit{endDistanceToGoal}, \textit{outcome}, \textit{distanceToPost}, and \textit{distanceToBar}. These features include information about the outcome of actions. They inform the algorithm on the outcome of the action, which impacts how the action is valued. By introducing these features, our model evaluates the execution of the action by the player. When these features are not included the model measures a player's intent. We call the models Outcome-aware VAEP (O-VAEP) and Intent VAEP (I-VAEP).

In opposition to the original proposal \cite{decroos_actions_2019}, we build a regression model. Instead of splitting the state value between defensive and offensive values, we use a continuous label between -1 and 1, where -1 indicates certainty that the team will concede, and 1 indicates certainty that the team will score a goal.

We describe the label’s value $l$ in Equation \ref{eq:label}, where $label_e$ indicates the value of the label of event $e$, $T$ is the team performing the event, $t$ is the time of the event in minutes, and $O$ is the ordinal number of the action (after sorted by time of occurrence). The component $\max\{1 - (t_{e} - t_{goal}), 0\}$ relates to how much time it occurs before the action, capped at 1 minute, and $int(O_e - O_{goal} > 5)$ relates to whether the actions is one of the last 5 actions before a goal. The component $(2*(T_e == T_{goal}) - 1)$ relates to whether the team that scored is the same team that performed the event and gives negative values to actions that lead to conceding goals. After calculating the label, we check whether the action occurred in the same period/game to ensure data consistency.

\begin{equation}
l_e=(2*(T_e == T_{goal}) - 1) * \max \{ \max\{1 - (t_{e} - t_{goal}), 0\}, int(O_e - O_{goal} > 5)\}
\label{eq:label}
\end{equation}

To train our model, we use the \textit{Random Forest} learner from \textit{scikit-learn} default parameters, except for \textit{min\_samples\_split}, which was set to 50 to avoid overfitting the model. Both models have the same parameters. We measure \textit{Mean Absolute Error} (MAE), and \textit{Median Absolute Error} (MedAE) to evaluate our model, and the results are available on the Appendix \ref{ap:results}.

Having estimated the probability of scoring/conceding at the end of each event, we can now calculate how much value an action creates. For this, we calculate the difference between the current probability of scoring and probability two actions earlier. We opted for this approach since our dataset contains many paired actions. For example, if a foul precedes a dribble, both actions should contribute in the same magnitude to the probability of scoring. Using a lag of two, we ensure we adequately value paired actions. To ensure consistency, we remove kick-off actions.

\section{Creating tracking metrics} \label{sec:Creating tracking metrics}
After calculating the value of each action, there is one additional step to build the tracking metric: grouping all player actions per game, building a time series that contains a player's total VAEP in each game. To ensure consistency in data, we calculate VAEP per minute played and only kept games where the player played more than 60 minutes.

Besides tracking all actions, we can track a specific subset of actions. This paper also generates time series related to passes, dribbles, and shots. Due to the specificity of the position, we excluded goalkeepers from the analysis.

\begin{figure}
\includegraphics[width=\textwidth]{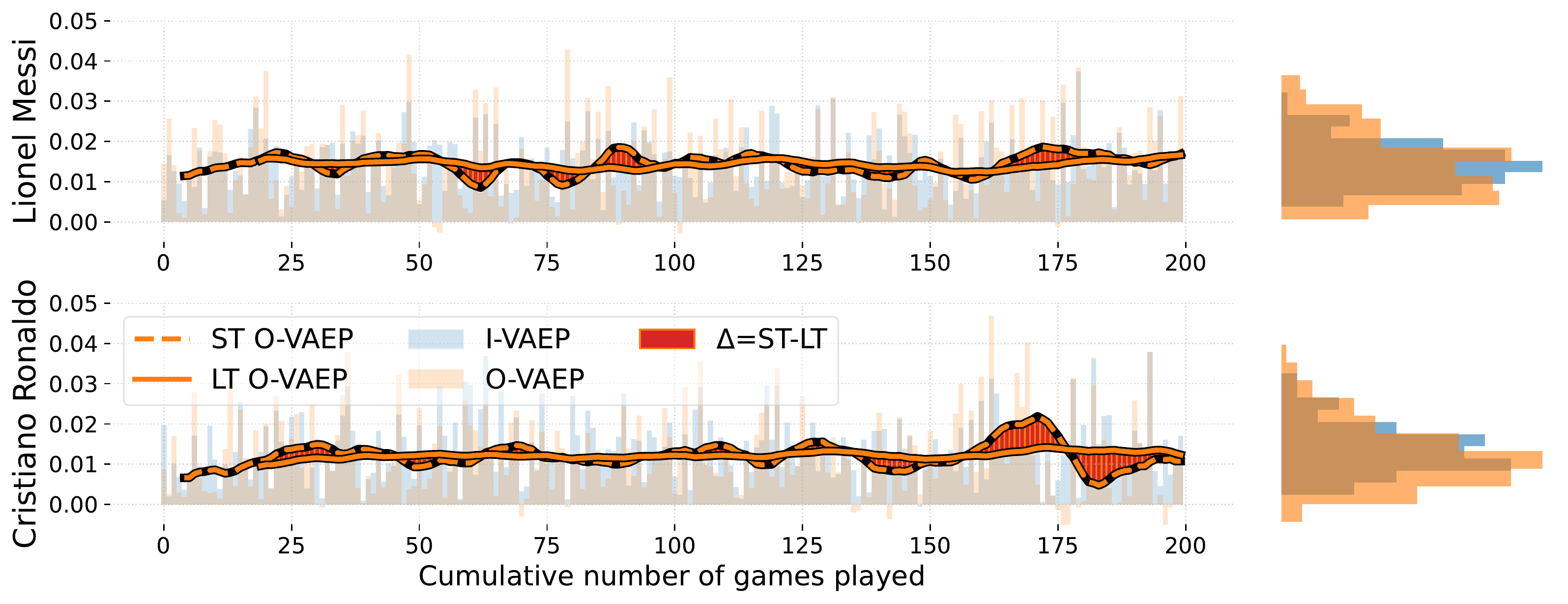}
\caption{On the left, we present the I-VAEP and O-VAEP values per game in the background, and the short-term and long-term ratings with the difference between both ratings in front. On the right, we observe the player's game ratings distribution.} \label{fig:dataviz}
\end{figure}

As shown in Figure \ref{fig:dataviz}, performance per game is very volatile. To extract meaningful information, we need to capture the trend within the time series, ignoring the fluctuations between games. We use moving averages across different windows to capture the trend in our data. Another option would be using exponential moving averages, but our experiments showed that they are less robust to outliers.

Before calculating the moving averages, we need to define the moving windows. The first decision is which index we will use in our series. We have two options: (1) use the cumulative count of games played, or (2) use game time. In the first option, we would define the window as a number of games. The series will have consistent indexes without problems regarding missing data from injuries or season breaks. On the second option, we define the window as the number of days. The index is inconsistent, but this format contains more information, like periods of player injury or players leaving the league. We use a game-count index for this work, due to its simplicity.

To track the evolution of players, we create two different metrics: short-term and long-term player performance. Each metric has a specific role. The short-term metric captures the short-term trend of a player without overreacting to a single performance, acting as a proxy for a player’s recent form. The long-term metric considers the player’s long-term performance and provides a consistent, less volatile measure of player performance.

In this work, we set the short-term window to 10 games, with a minimum of 5 games in this period. The long-term window is 40 games with at least 20 games. We set these values by trial and error in the training set. Our goal was to create a metric that captured player quality as fast as possible without overreacting to one game's performances.


\section{Experimental Use Cases} \label{sec:Experimental Use Cases}
\subsection{Evaluating Players}

\begin{figure}
\includegraphics[width=\textwidth]{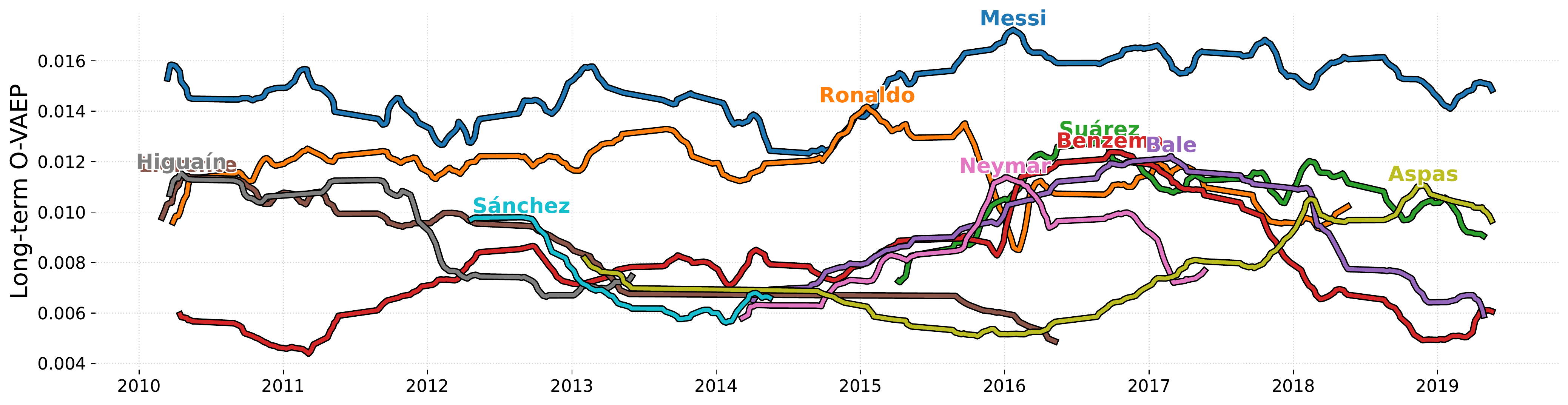}
\caption{The long-term O-VAEP ratings of the 10 players with highest peak.} \label{fig:ltovaep}
\end{figure}

The first use case for our performance metrics is to evaluate the best players in the league. Figure \ref{fig:ltovaep} shows the long-term O-VAEP rating over time.

We observe that Lionel Messi was the best player in La Liga during the period in our dataset. Except for a small period at the end of 2014, when Cristiano Ronaldo took the lead, Messi dominated the first place. Ronaldo is also a clear second most of the time, until he is overtaken by several players in 2016. Luis Suárez closes the podium. It is interesting to note that many series look correlated, even for players of different teams, which might indicate that competition between top players might get the best out of them.



\subsection{Rating's Granularity}

\begin{figure}
\includegraphics[width=\textwidth]{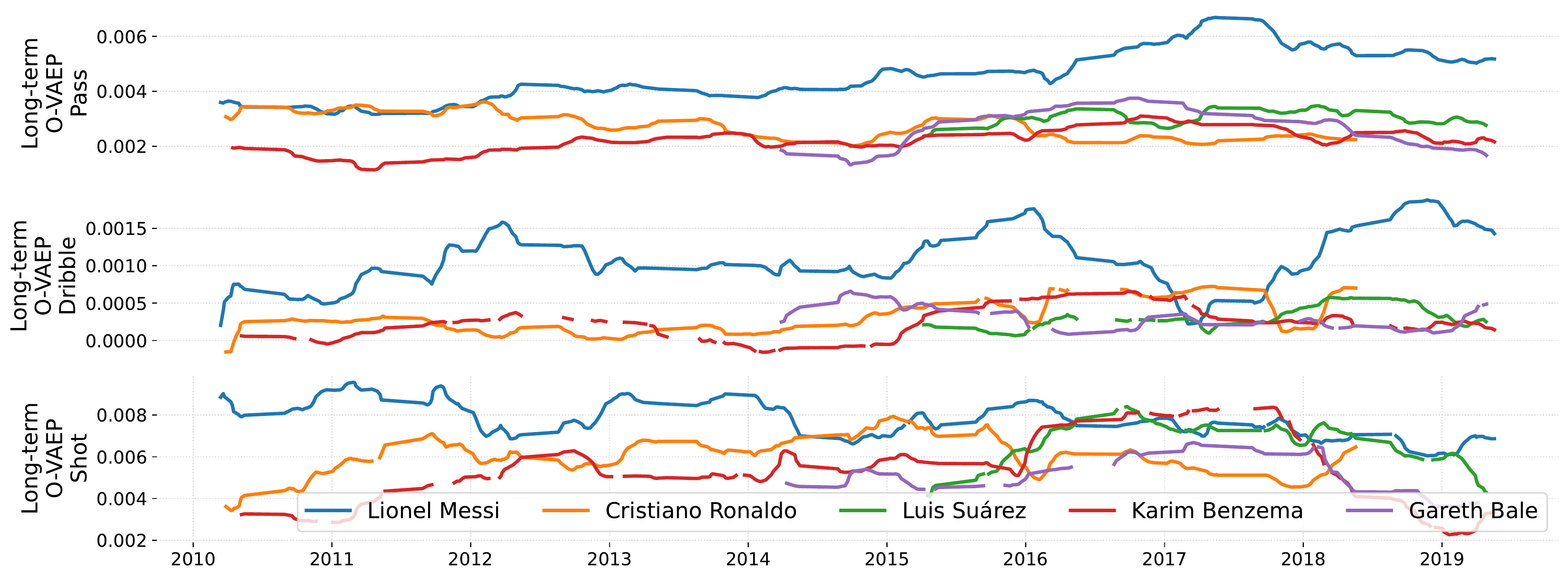}
\caption{The contribution of different action types to the player's O-VAEP ratings.} \label{fig:granular}
\end{figure}

We can also evaluate players' ratings by their performance in specific actions. Figure \ref{fig:granular} presents the VAEP ratings of passing, dribbling, and shooting actions over time for the five players with the highest OA-VAEP values.

Unsurprisingly, Messi leads in value created from passes and dribbles for nearly the entirety of the data set. But more interesting than analyzing who is best at what is to understand player changes over time. In Figure \ref{fig:granular}, we observe that the value Messi creates from dribbles follows the opposite behavior of the value he creates from passes. As Barcelona’s key playmakers like Xavi and Iniesta retired, Messi adapted his game, increasing his role in creating changes through passes and sacrificing the value he can produce by dribbling.

\subsection{Player Consistency}

Consistency is a key factor for player selection. A consistent player provides a similar amount of value to the team across all games. Especially when contending for championships, exhibition consistency is key to offering guarantees that teams do not miss a seasonal objective due to a single inconsistent match. 

Equations \ref{eq:gametogamevol}, \ref{eq:negativegamevol}, and \ref{eq:negativeshorttermvol} describe our volatility metrics for player performance, where $\vec{r_{G}}$ is vector of a player’s game rating, $\vec{r_{ST}}$ is vector of a player’s short-term rating, $\vec{r_{LT}}$ is vector of a player’s long-term rating for a specific game, and $\sigma$ is the standard deviation function.

\begin{equation}
Game\ to\ Game\ Volatility = \sigma(\vec{r_{G}} - \vec{r_{LT}})\label{eq:gametogamevol}
\end{equation}

\begin{equation}
Negative\ Game\ Volatility = \sigma(\Delta * (\Delta < 0)),\ where\ \Delta = \vec{r_{G}} - \vec{r_{LT}}
\label{eq:negativegamevol}
\end{equation}

\begin{equation}
Negative\ Short\ Term\ Volatility = \sigma(\Delta * (\Delta < 0)), where\ \Delta = \vec{r_{ST}} - \vec{r_{LT}}
\label{eq:negativeshorttermvol}
\end{equation}

By measuring the standard deviation of the difference between a player’s exhibitions and his long-term performance, we can obtain a proxy for how consistent a player is in delivering his average performance. 

\begin{figure}
\includegraphics[width=\textwidth]{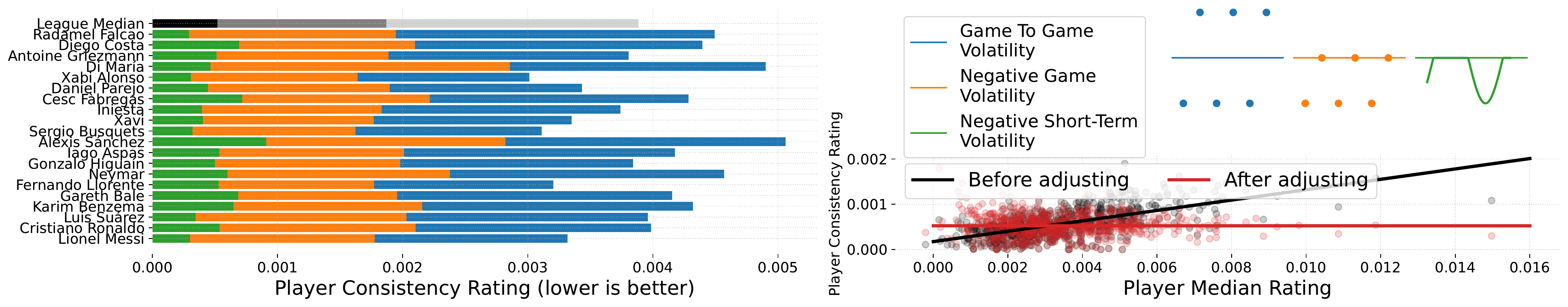}
\caption{On the left: the volatility ratings of the top 10 players in our data set, plus 10 additional players arbitrarily selected. On the right: a visual explanation of the ratings on top, and the process of adjusting volatility according to the median rating of the player on the bottom.} \label{fig:consistency}
\end{figure}

However, due to the nature of our ratings, higher-rated players will have higher standard deviations. To fix this, we fit a linear regression model and use it to control for player ratings. Figure \ref{fig:consistency} presents the transformation of the volatility metric and shows the consistency results for the top players in our dataset.

Although we did not explore this, it would be interesting to evaluate a player’s consistency across different ranges of game difficulty. For example, one could find a set of players that play better under challenging games and another that exceeds in more accessible games. This can be key when formulating the squad since the coach will want options for every type of opposition the team will face.

\subsection{Player Development Curve}
With access to information about players' performance over time, we can understand how a player develops on average. Figure \ref{fig:pdc} shows the player development curve (PDC) obtained from the procedure described in Appendix \ref{ap:pdc}.

\begin{figure}
\includegraphics[width=\textwidth]{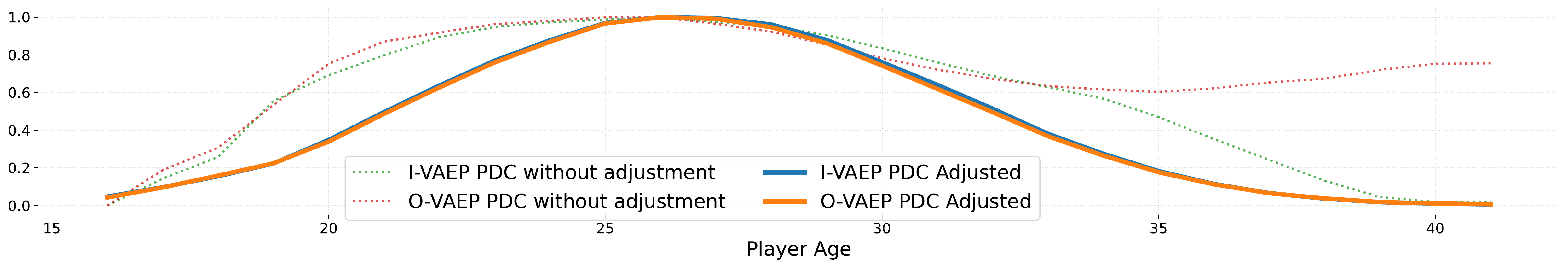}
\caption{The average PDC, from the player's lowest (0) to the highest rating (1).} \label{fig:pdc}
\end{figure}

To build the PDC, we group the player's ratings relative to their peak across all ages available in the dataset. However, we face a bias in our dataset after performing this step. There are much fewer young (>20 years old) and older (>33 years old) players in our dataset. This occurs because players in these age groups will only be playing in the first team if they are exceptional.

This means that the average young/old player in our dataset is not the same compared to the average of a uniform sample of young/old players, who are more likely to play in academies or lower divisions. The dataset contains a sample of the top young/old players. To control for this, we assume that the number of players across all ages should be uniform. Therefore, we multiply the value of each age by $1-relative\ amount\ of\ players$.

\begin{figure}
\includegraphics[width=\textwidth]{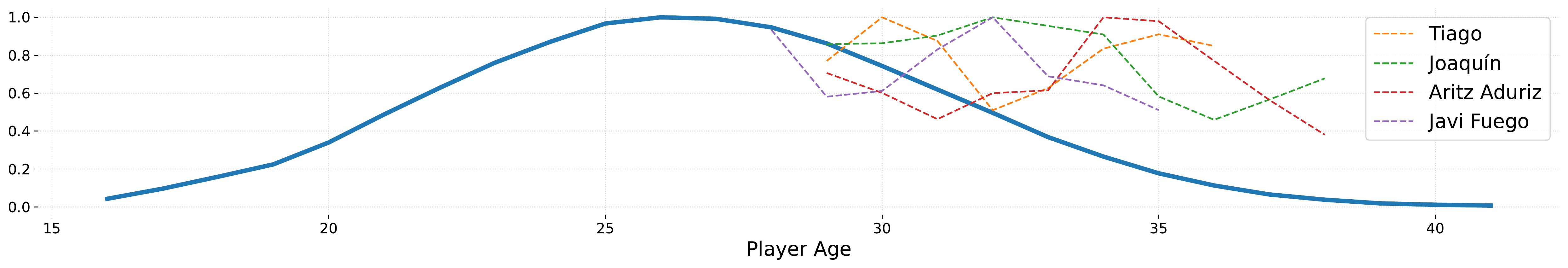}
\caption{A sample of late bloomers in our dataset. These players either hit or were close to hitting their peak performance much later than the average player.} \label{fig:latebloomers}
\end{figure}

Agreeing with similar studies in the subject \cite{dendir_soccer_peak_2016}, the PDC shows players peaking between 25 and 27 years old. The curve also relates the equivalent age where a young player starts to provide more value to the team compared to an older player. 
One interesting application of the PDC is to detect late bloomers. Late bloomers are players who reach their peak long past the 25 to 27 years old range. Since the players’ market value decreases substantially after their peak age, clubs can find bargains by purchasing older players who still have room to improve. This is the case with players presented in Figure \ref{fig:latebloomers} like Tiago, Aritz Aduriz, and Joaquín, who played at their peaks in their mid-30s.

\section{Discussion} \label{sec:Discussion}
This work argued about potential use cases for player ratings treated as time series. The main goal was to highlight some situations where handling ratings as a continuous evaluation mechanism can yield better results and novel methods when compared to handling the problem as a tabular task.

In player valuation, we showed that tracking a rating over time can help us identify clear trends in player performance. Furthermore, it allows for a comparison between players over time. Besides ratings, we can measure players’ consistency, discriminate which actions the players use to provide value and understand long-term player style changes over time.

Our approach does not come without limitations: our methods demand many data points for the ratings to be reliable, which might not always be available. Although we can evaluate the most important players reliably, we cannot assign ratings to players with smaller game samples. We cannot simply reduce the number of games required: doing so would result in unreliable and unstable ratings. We require methods to rate players that do not rely on so many data points. Another drawdown is the lack of cap on how old a game can be and still count for a player's rating. Due to our game-count index, players keep their rating even after years without playing a game.

Event data contains much information but has limitations. The context in certain actions (like tackles and interceptions) is limited, and the models wrongly value defensive actions. That being said, we still see many avenues where we can improve our results. For example, normalizing player ratings by position, game difficulty, competition difficulty, and teammates ratings are some of the possibilities where we can reduce biases in our current framework.

The PDC shows some of the potentials of treating data as time series. At this moment, we defined a curve that shows the expected player development over different ages. However, we see different use cases that can help in many facets of a team’s management. Locating where a player is within the curve, detecting late bloomers, or projecting future player performance are some of the possibilities where this method can generate value for a club.

\section{Conclusion} \label{sec:Conclusion}
This paper presented a framework to treat player ratings as time series. First, we defined our VAEP models and methodology to measure short-term and long-term player performance. Then, we presented several use cases for the time series produced. From player evaluation to player development, we see the potential to apply this approach in many parts of the game.

As for future work, we plan to address the following:
\begin{itemize}
    \item Adding external context to the time series (such as game difficulty);
    \item Understand the trade-off between average value versus total value per game;
    \item Improve the quality of the valuation of defensive actions;
    \item Increase the efficiency of data, decreasing the amount of data required for ratings without decreasing its quality.
\end{itemize}





\subsubsection{Acknowledgements}
This work is financed by National Funds through the Portuguese funding agency, FCT - Fundação para a Ciência e a Tecnologia, within project UIDB/50014/2020.

The second author would like to thank Futebol Clube Porto – Futebol SAD for the funding.
Any opinions, findings, and conclusions or recommendations expressed in this material are those of the authors and do not necessarily reflect the views of the cited entity. 

%
%
%
%





\bibliographystyle{splncs04}
\bibliography{samplepaper}

\newpage

\appendix
\section{Complementary Information for Reproducibility}
\subsection{VAEP Models Feature List} \label{ap:features}
The feature set of our VAEP models had the following features for each action:
\begin{itemize}
    \item \textbf{period} - which half of the game the action took place in.
    \item \textbf{second} - second of the action, relative to the start of the half.
    \item \textbf{x}, \textbf{endX} - position of the ball on the long axis (x-axis).
    \item \textbf{bodyPart} - whether the action was performed with the head.
    \item \textbf{encType} - the encoded value of the action type.
    \item \textbf{startAngleToGoal}, \textbf{startDistanceToGoal}, \textbf{endAngleToGoal}\footnotemark[1], \textbf{endDistanceToGoal}\footnotemark[1] - distance and angle from the ball position to the opposition’s goal at the start/end of an action.
    \item \textbf{intentProgressive}\footnotemark[2] - whether a pass ended closer to the goal than the starting position.
    \item \textbf{encPreviousActionType1}, \textbf{previousActionTeamIsEqual1}, \textbf{encPreviousActionType2}, \textbf{previousActionTeamIsEqual2} - information about the previous actions.
    \item \textbf{outcome}\footnotemark[1], \textbf{distanceToPost}\footnotemark[1], \textbf{distanceToBar}\footnotemark[1] - variables related to the outcome of actions. *outcome* indicates whether the action was successful, while the distance metrics refer to the placement of the shot in the goal mouth and are used to measure how well a shot was executed. The outcome of shot actions is always set to True.
\end{itemize}
\footnotetext[1]{Only for the O-VAEP model.} 
\footnotetext[2]{Only for the I-VAEP model.} 

\newpage
\subsection{Comparison Between Labeling Strategies} \label{ap:results}

\begin{figure}
\centering
\includegraphics[trim=10cm 0cm 0cm 6cm, clip, width=0.8\textwidth]{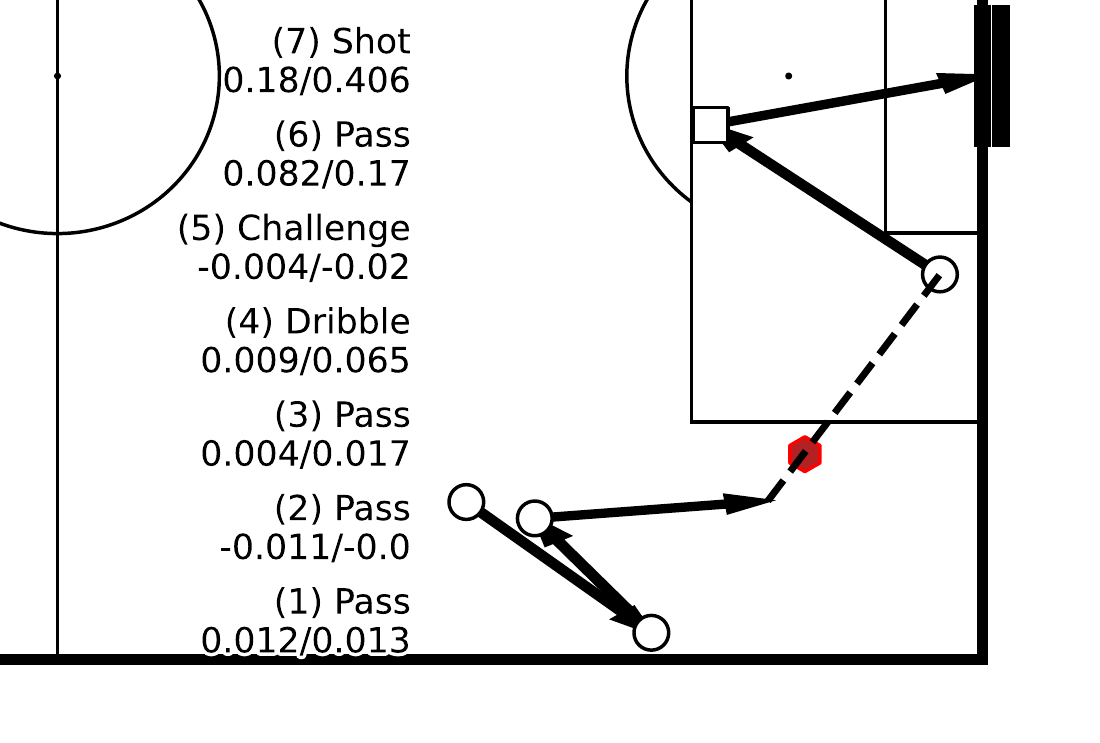}
\caption{Our results on the same play (Barcelona (3)-0 Real Madrid, December 23rd, 2017) presented by Decroos et al. \cite{decroos_actions_2019}. Results represent I-VAEP/O-VAEP, respectively.} \label{fig:originalplay}
\end{figure}

This section presents the Table \ref{tab:Algorithm Performance} comparing the algorithm's performance against using a similar labeling strategy to Decroos et al. \cite{decroos_actions_2019}. Note that although the results seem to favor our approach, they only benchmark both labeling approaches. In Figure \ref{fig:originalplay} we show our results for comparison with Decroos et al. \cite{decroos_actions_2019}

\begin{table}
\label{tab:Algorithm Performance}
\caption{Model Performance}
\centering
\begin{tabular}{|l|l|l|l|l|l|l|}
\hline
Model & Dataset & Label Type & MAE & Change \% & MedAE & Change \%\\
\hline
I-VAEP & Train & k=10 label & 0.02681 & - &0.00617 & -\\
I-VAEP & Train & Our label & 0.02748 & Up 2.5\% & 0.00585 & Down 5.2\%\\
O-VAEP & Train & k=10 label & 0.02658 & - & 0.00608 & -\\
O-VAEP & Train & Our label & 0.02622 & Down 1.4\% & 0.00566 & Down 6.9\%\\
I-VAEP & Test & k=10 label & 0.03259 & - & 0.00758 & -\\
I-VAEP & Test & Our label & 0.03266 & Up 0.2\%  & 0.00713 & Down 5.9\%\\
O-VAEP & Test & k=10 label & 0.03233 & - & 0.00752 & -\\
O-VAEP & Test & Our label & 0.03127 & Down 3.3\% & 0.00693 & Down 7.8\%\\
\hline
\end{tabular}
\end{table}

\newpage
\subsection{Building the Player Development Curve} \label{ap:pdc}
\begin{figure}
\centering
\includegraphics[width=\textwidth]{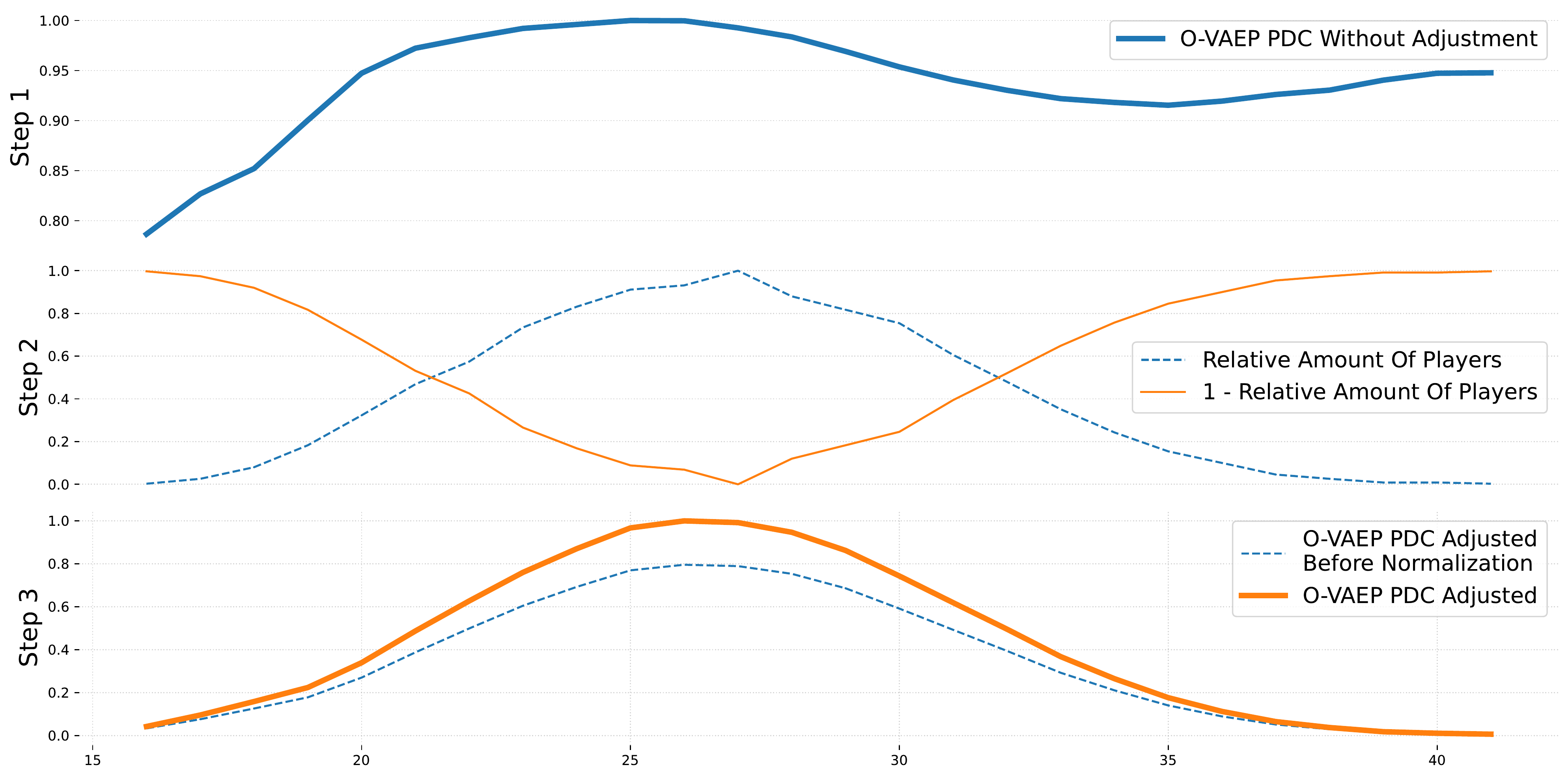}
\caption{Obtaining the player development, step-by-step.} \label{fig:pdc_step_by_step}
\end{figure}

To build the PDC, we perform the following steps:

\begin{itemize}
    \item For every player
    \begin{itemize}
        \item Check if the player has played for more than one season.
        \item Calculate the player's median performance rating grouped by age.
        \item Normalize the player age-performance curve by dividing every value by the maximum.
    \end{itemize}
    \item Calculate the median performance rating for every age available across all players, the unadjusted player development curve (See Figure \ref{fig:pdc_step_by_step}, Step 1).
    \item Calculate the relative amount of players, which is the amount of players at a certain age divided by the maximum amount of players at any age (See Figure \ref{fig:pdc_step_by_step}, Step 2).
    \item Adjust the player development curve by multiplying the value at each age by $(1 - relative\ amount\ of\ players)$ (See Figure \ref{fig:pdc_step_by_step}, Step 2).
    \item Smooth the obtained curve and normalize between 0 and 1  (See Figure \ref{fig:pdc_step_by_step}, Step 3).
\end{itemize}

\end{document}